\documentclass[sigconf]{acmart}
\usepackage[table,xcdraw]{xcolor}
\usepackage{graphicx}
\usepackage{float}
\usepackage{pdfpages}

\AtBeginDocument{%
  }

\setcopyright{acmlicensed}
\copyrightyear{2026}
\acmYear{2026}
\acmDOI{XXXXXXX.XXXXXXX}

\acmConference[ICMI '26]{28th International Conference on Multimodal Interaction}{October 5--9, 2026}{Napoli, Italy}
\acmISBN{978-1-4503-XXXX-X/2026/10}

\begin{document}
\pagestyle{plain}

\title{FAIR\_XAI: Improving Multimodal Foundation Model Fairness via Explainability for Wellbeing Assessment}

\newcommand{\equalcontrib}{\authornote{Equal contribution.}}
\newcommand{\corr}{\authornote{Corresponding author. J. Cheong is supported by a DAAD AInet Fellowship.}}
\newcommand{\ackn}{\authornote{The works of F. I. Do\u{g}an \& H. Gunes were supported in part by CHANSE \& NORFACE through the MICRO project, funded by ESRC/UKRI grant ref.~UKRI572.}}

\author{Sophie Chiang}
\orcid{0000-0002-9249-2019}
\authornote{Equal contribution.}
\email{yhc49@cam.ac.uk}
\affiliation{%
  \institution{University of Cambridge}
  \country{UK}
}
\author{Tom Brennan}
\orcid{0009-0001-1686-0342}
\authornotemark[1]
\email{tb800@cam.ac.uk}
\affiliation{%
  \institution{University of Cambridge}
  \country{UK}
} 

\author{Fethiye Irmak Doğan}
\ackn
\orcid{0000-0002-1733-7019}
\email{fid21@cam.ac.uk}
\affiliation{%
  \institution{University of Cambridge}
    \country{UK}
}
\author{Jiaee Cheong} 
\corr
\orcid{0000-0001-5964-2284}
\email{jc2208@cam.ac.uk}
\affiliation{%
  \institution{University of Cambridge,\\ Harvard University}
  \country{UK, US.}
}

\author{Hatice Gunes}
\authornotemark[2]
\orcid{0000-0003-2407-3012}
\email{hatice.gunes@cl.cam.ac.uk}
\affiliation{%
  \institution{University of Cambridge}
  \country{UK}
}

\begin{abstract}
In recent years, the integration of multimodal machine learning in wellbeing assessment has offered transformative potential for monitoring mental health. However, with the rapid advancement of Vision-Language Models (VLMs), their deployment in clinical settings has raised concerns due to their lack of transparency and potential for bias. 
While previous research has explored the intersection of fairness and Explainable AI (XAI), its application to VLMs for wellbeing assessment and depression prediction remains under-explored. 
This work investigates VLM performance across laboratory (AFAR-BSFT) and naturalistic (E-DAIC) datasets, focusing on diagnostic reliability and demographic fairness. 
Performance varied substantially across environments and architectures; Phi-3.5-Vision achieved 80.4\% accuracy on E-DAIC, while Qwen2-VL struggled at 33.9\%.
%
%
Although bias existed across both architectures, Qwen2-VL showed higher gender disparities, while Phi-3.5-Vision exhibited more racial bias.
Our XAI intervention framework yielded mixed results; fairness prompting achieved perfect equal opportunity for Qwen2-VL at a severe accuracy cost on E-DAIC, while an attribution-guided training penalty collapsed to a single class predictor.
On AFAR-BSFT, explainability-based interventions improved procedural consistency 
but did not guarantee outcome fairness, sometimes amplifying racial bias.
These results highlight a persistent gap between procedural transparency and equitable outcomes. We analyse these findings and consolidate concrete recommendations for addressing them, 
emphasising that future fairness interventions must jointly optimise predictive accuracy, demographic parity, and cross-domain generalisation.

%
\end{abstract}



\maketitle

\section{Introduction}
Clinical depression remains one of the most prevalent mental health disorders, affecting approximately 332 million people worldwide, and is a leading cause of disability \cite{who2025depression,chen2026current}.
The application of multi-modal machine learning (ML) to wellbeing assessment has seen a significant surge in popularity, offering promising methods for monitoring mental health illnesses and predicting patient outcomes \cite{song2025automatic,li2025automated}.
More recently, this trend has accelerated further with the adoption of Multimodal Foundation Models, and, more specifically, Vision-Language Models (VLMs)~\cite{lian2025mer,dwyer2025mindbench}.
Unlike traditional supervised learning, VLMs are often pre-trained on large-scale image-text corpora \cite{dogan2026investigating,zhang2024vision}, allowing for zero-shot reasoning and prediction without explicit re-training~\cite{radford2021learning,abbasi2025robot,flathers2025interpreting,cheong2026automated}, opening the door to more scalable and autonomous clinical solutions.

Despite this potential, the deployment of VLMs in clinical settings is currently bottlenecked by a ``black-box" dilemma \cite{ratti2022explainable}.
%
This lack of transparency can make it difficult for clinicians to discern whether a classification is based on valid clinical markers, or a result of ML prediction bias~\cite{
cheong2021hitchhiker,chuang2023debiasing, abbasi2025robot,cheong2026fairssl}, i.e., when there is prejudice or favouritism towards an individual or group based on their inherent or acquired characteristics~\cite{mehrabi2021survey}.
%
%
Explainable AI (XAI) has been proposed to bridge this gap between performance and trust.
XAI refers to the suite of ML techniques that produce more explainable models while maintaining a high level of performance, and enable humans to understand and effectively manage the emerging generation of artificially intelligent partners~\cite{arrieta2020explainable}.
%
%
However, while previous studies have explored the intersection of fairness and explainability~\cite{zhao2023fairness, dogan2026investigating}, the relationship for VLMs in wellbeing assessment, and more specifically depression classification, is significantly under-explored.

This paper therefore investigates the following research questions:
\textbf{\textit{(RQ1)}} How does zero-shot VLM depression classification performance vary between controlled laboratory datasets and naturalistic, unconstrained clinical environments?
It is important to understand how VLM performance can vary across different types of data domains, and to determine how environmental noise and data distribution shifts impact VLM diagnostic reliability.
We also ask \textbf{\textit{(RQ2)}}: To what extent do systematic performance disparities (e.g. gender-based accuracy) persist across VLM architectures, and are these biases consistent across both controlled and naturalistic data domains?
Since we use multiple VLM architectures, we can evaluate whether any bias that exists is model or domain-specific.
Finally, we ask \textbf{\textit{(RQ3)}}: Can an XAI-guided intervention method be systematically deployed to improve VLM outcome fairness?

The contributions of this paper are therefore threefold.
First, we provide a cross-domain benchmark demonstrating how zero-shot VLM performance in depression classification shifts between naturalistic and laboratory settings.
Our findings revealed that performance varied substantially across environments and architectures.
Second, we present a fairness audit that reveals how and what biases are model or domain-specific.
Finally, we audit the promise and the limits of our XAI-guided intervention, which yielded mixed results; an increase in procedural consistency did not always translate to equitable outcomes, and in some cases, even amplified bias, or caused model collapse to a single class.
%
%
We collate the resulting insights into actionable recommendations (Table~\ref{tab:recommendations}) for the responsible deployment of VLMs in clinical wellbeing assessment.

\section{Related Work}

\textbf{Multimodal Wellbeing Assessment.}
The integration of multimodal data, 
has become increasingly standard 
as it captures a richer set of human behaviours than traditional unimodal approaches~\cite{poria2017review,cheong2024fairrefuse,cheong2025u,kwok2025machine}.
Multimodal data has been used in various affective computing tasks, such as depression assessment~\cite{alghowinem2016multimodal,cheong2024fairrefuse,gui2019cooperative,cheong2025fairwell} and automated pain detection~\cite{aung2015automatic,green2025gender}.
Recently, this field has seen a significant shift from traditional statistical models to large pre-trained foundation models.
Initially, this was led by Large Language Models (LLMs) for text-based mental health support~\cite{laban2024lexi,spitale2024underneath}, which also later spurred the development of conversational AI agents~\cite{dong2023towards}, as well as the use of social robotics for stress reduction~\cite{laban2025people,spitale2025exploring}.
Concurrently, VLMs have emerged and have been noticed for their zero-shot and few-shot capabilities, as well as their use for such multimodal tasks. 
For instance, Abbasi \textit{et al.} demonstrated the use of a lightweight, open-source VLM in a robot-led approach for child wellbeing assessment~\cite{abbasi2025robot}.
%
%

\vspace{1mm}
\noindent
\textbf{Fairness in Affective Computing.}
Bias in affective computing has been well-documented~\cite{cheong2023causal,cameron2024multimodal,cheong2025small}, 
especially regarding gender~\cite{green2025gender,cheong_gender_fairness}
and ethnicity \cite{cameron2024multimodal,cheong2022counterfactual}.
%
%
Domnich \textit{et al.}~\cite{domnich2021responsible} investigated gender fairness of ML models in emotion recognition, and Cameron \textit{et al.}~\cite{cameron2024multimodal} for pain detection from facial expressions.
%
%
Lee \textit{et al.}~\cite{lee2021included} not only found performance gaps between gender and ethnicity groups, but also gender-correlated features that help explain why this is the case.
Recent LLM~\cite{spitale2024underneath} and VLM~\cite{abbasi2025robot,dogan2026investigating} studies have mirrored these concerns.
There is also a concern that these models may exaggerate existing healthcare biases, due to the social prejudice already present in training data~\cite{guo2024bias,taubenfeld2024systematic,cheong2026equitable}.
While these studies mainly focus on auditing the existence of bias in model outcomes, our work aims to extend this by investigating how we can use XAI-driven insights to actively mitigate these disparities in VLMs without task-specific fine-tuning.

\vspace{1mm}
\noindent
\textbf{Explainable AI.}
%
%
Previous research has developed XAI methods that have been shown to be important in debugging~\cite{nori2019interpretml,dwivedi2023explainable, cortinas2023toward}, building trust~\cite{thalpage2023unlocking}, and diagnosing bias in various prediction tasks~\cite{arias2022focus}.
Helbling \textit{et al.} introduced the idea of a Concept Attention framework, which offers higher-level interpretability 
to visualise how a model links semantic prompts to visual elements~\cite{helbling2025conceptattention}.
%
%
Recent studies have started to explore the intersection of fairness and explainability.
%
For example, Lundberg \textit{et al.} demonstrated how model disparity can be decomposed into feature-specific contributions, in order to locate the source of bias~\cite{lundberg2020explaining}.
Furthermore, Zhao \textit{et al.}~\cite{zhao2023fairness} presented a set of metrics (Ratio and Value-based Explanation Fairness) in order to mitigate procedure-based bias, in addition to a Comprehensive Fairness Algorithm that improved traditional fairness, satisfied explanation fairness and maintained utility performance.
Despite these advancements, the intersection between fairness and explainability remains under-explored for large-scale VLMs in affective computing.
%

\section{Methodology}
\begin{figure*}[t]
  \centering
  \includegraphics[width=\textwidth]{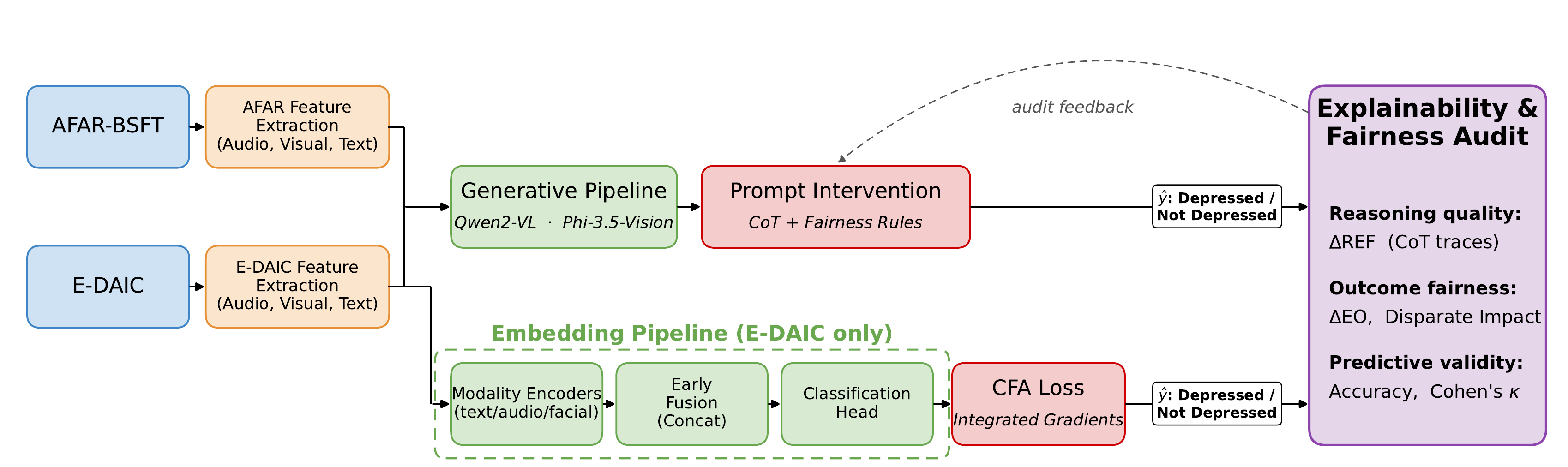}
  \caption{FAIR\_XAI framework. AFAR-BSFT (lab) and E-DAIC (in-the-wild) feed per-modality feature extraction into a \textbf{generative pipeline} (zero-shot Qwen2-VL / Phi-3.5-Vision under a CoT + fairness-rule intervention) and, for E-DAIC, an \textbf{embedding pipeline} fusing frozen-VLM text and MLP audio/facial embeddings under a counterfactual-fairness (CFA) loss with integrated-gradient attributions. Both pipelines emit binary depression predictions ($\hat{y}$) into a shared audit ($\Delta REF$, $\Delta EO$, disparate impact, accuracy); audit findings feed back into the intervention (dashed).}
  \Description{Figure 1 shows two datasets, AFAR-BSFT and E-DAIC, each feeding into a per-modality feature extraction stage. Both datasets feed into a shared generative classification pipeline using zero-shot VLMs and a prompt-based intervention. E-DAIC additionally feeds into an embedding pipeline with modality encoders, early fusion, a classification head, and a counterfactual fairness algorithm. Outputs from both pipelines are routed into an explainability and fairness audit block, which loops back into the intervention design.}
  \label{fig:overall_framework}
\end{figure*}

The overall pipeline is illustrated in Figure~\ref{fig:overall_framework}.
First, we have per-modality feature extraction from raw audio, visual and verbal signals.
Second, two parallel classification pathways, a zero-shot \emph{generative} pipeline in which the VLM is prompted directly and an \emph{embedding}-based fusion pipeline in which frozen VLM text encodings are combined with audio and facial MLP encodings.
Third, XAI-guided interventions via prompt-based chain-of-thought and fairness rule set for the generative pathway and a counterfactual fairness loss for the embedding pathway.
Fourth, a joint explainability-and-fairness audit that scores reasoning quality ($\Delta REF$) alongside outcome fairness ($\Delta EO$, disparate impact, accuracy gaps). 

\subsection{Problem Formulation}
The primary task is defined as binary depression classification.
For a given dataset $\mathcal{D} = \{(x_i, y_i )\}^n_{i=1}$ where $i \in \mathcal{I}$ represents the participant index, the goal is to predict a label $\hat{y_i} \in \mathcal{Y}$ from input feature vector $\mathbf{x}_i \in \mathcal{X}$.
The label space is defined such that $\mathcal{Y} = \{0,1\}$, where $y_i=  0$ denotes a healthy non-depressed individual and $y_i=1$ indicates a clinical depression diagnosis.
For each participant $i$, the input $\mathbf{x}_i$ is a composite multimodal representation: $\mathbf{x}_i = (x_i^{\text{text}}, x_i^{\text{audio}}, x_i^{\text{face}})$, where $x_i^{\text{text}}$ is the interview transcript, $x_i^{\text{audio}} \in \mathbb{R}^{d_a}$ contains OpenSMILE eGeMAPS acoustic descriptors, and $x_i^{\text{face}} \in \mathbb{R}^{d_f}$ contains OpenFace~2.0 Action Unit, gaze and head-pose features.

Our generative pipeline (Section~\ref{sec:generative_pipeline}) retains $\mathbf{x}_i$ in its native modality space and fuses modalities via string concatenation, serialising a mutually-informative subset of $\mathbf{x}_i$ together with demographic metadata and a task instruction into a single natural-language prompt.
Meanwhile, our embedding pipeline (Section~\ref{sec:embedding_pipeline}) projects each modality independently into a fixed-dimensional embedding ($\mathbf{h}_{\text{text}} \in \mathbb{R}^{256}$ and $\mathbf{h}_{\text{audio}}, \mathbf{h}_{\text{face}} \in \mathbb{R}^{128}$), and fuses them by early concatenation into $\mathbf{h}_{\text{fused}} \in \mathbb{R}^{512}$, which is passed to a binary classification head.
In both pipelines the VLM reasons over pre-extracted, standardised descriptors (OpenSMILE eGeMAPS, OpenFace Action Units) rendered as text rather than over raw pixels or audio. This is deliberate: our aim is not to benchmark visual perception but to audit fairness and explainability under a controlled reasoning setting, where every model receives the same structured, visually-derived evidence and perception noise cannot confound reasoning-level bias. Gao \textbackslash{}textit\{\textit{et al.}\}~\textbackslash{}cite\{gao2025moma\} adopts a similar textualised-input design for clinical prediction with a VLM.

\subsection{Generative Classification Pipeline}
\label{sec:generative_pipeline}
Our generative classification pipeline adopts a zero-shot prompting approach rather than full fine-tuning, avoiding the risk of over-fitting and allowing us to audit the inherent biases present in large-scale foundation models.

In order to maintain the prompt size within the context window limits for Qwen2-VL and Phi-3.5-Vision, we performed \textbf{feature selection} on the original 3000+ features.
This was done via calculating mutual information $I(x_j;y)$~\cite{shannon1948mathematical}, and sorting columns in descending order of features that would maximise the information gain between each feature $x_j$ and the prediction label $y$:

$$
I(X;Y) = \sum_{y \in Y} \sum_{x \in X} p(x,y) \log {\left( \frac{p(x,y)}{p(x)p(y)} \right)}
$$

We select the top $k$ features that maximised $I(x_j;y)$ to ensure the prompt contained the most statistically significant clinical biomarkers while remaining computationally efficient~\cite{amiri2011mutual,roubhi2025mutual}.
As neither pipeline updates model parameters, this selection leaves the generative pipeline zero-shot. It also shape which biomarkers the model sees, which we note as a limitation of the results.

For each participant $i$, the input prompt $\mathcal{P}_i$ was constructed via the concatenation of demographic metadata, selected clinical biomarkers, and a task-specific instruction set. This is formalised as follows:

$$
\mathcal{P}_i \leftarrow \mathcal{D}_i \oplus \mathcal{B}_i \oplus \mathcal{T}
$$

where $\mathcal{D}_i$ represents the string containing demographic information on participant $i$, such as gender and race.
Additionally, $\mathcal{B}_i = \{x_{i,1}, x_{i,2}, ..., x_{i,k}\}$ contains information regarding the selected biomarkers for participant $i$, and $\mathcal{T}$ represents the task instruction, including the expected output format of a score for the predicted label and brief reasoning.
Furthermore, $\oplus$ represents the string concatenation needed to form prompt $\mathcal{P}_i$, and the maximum token for the VLM output was set to 150 to allow reasoning to conclude.

\subsection{Embedding Classification Pipeline}
\label{sec:embedding_pipeline}
As an alternative to the generative classification pipeline, we introduce the embedding-based pipeline that extracts fixed-dimensional representations from frozen VLM encoders and trains a lightweight fusion classifier on top. This approach enables systematic ablation studies across modality combinations while amortising the computational cost of VLM inference.
By making our fusion classifier highly configurable, we compare performance across all combinations of text, audio and facial modalities; for each, we extract a fixed-dimensional embedding:

\begin{itemize}
    \item \textbf{Text}: We take interview transcripts through the VLM's text encoder and extract the final hidden state representation. For a given backbone model, e.g., Qwen2-VL-2B-Instruct (hidden size 1536) and Phi-3.5-Vision-Instruct (hidden size 3072), we applied a linear projection to 256 dimensions.
    
    \item \textbf{Audio}: OpenSMILE eGeMAPS features were processed by a 3-layer MLP to a 128-dimensional embedding.
    
    \item \textbf{Facial}: OpenFace 2.0 features were encoded via a 3-layer MLP to a 128-dimensional embedding.
\end{itemize}

Modality embeddings were then concatenated via early fusion: $\mathbf{h}_{\text{fused}} = [\mathbf{h}_{\text{text}}; \mathbf{h}_{\text{audio}}; \mathbf{h}_{\text{facial}}] \in \mathbb{R}^{512}$ (from our $256+128+128$ dimensions).
This fused representation is then given to a 2-layer classification head: $\mathbb{R}^{512} \xrightarrow{\text{Linear}} \mathbb{R}^{128} \xrightarrow{\text{ReLU+Dropout}} \mathbb{R}^{2}$, outputting logits for binary classification.
The fusion classifier and modality encoders (audio/facial MLPs) were trained end-to-end using cross-entropy loss with \textbf{class weights} to deal with the dataset's depressed/non-depressed class imbalance.

\subsection{Intervention Framework}
\label{sec:intervention}

\subsubsection{Prompt-Based Fairness Intervention} 
\label{sec:prompt_intervention}
For the generative classification pipeline (Section \ref{sec:generative_pipeline}), we use a lightweight prompt-based intervention that combines chain-of-thought (CoT) reasoning with explicit fairness instructions. Rather than modifying the data or retraining the model, we append two extra components to the task instruction $\mathcal{T}$ at inference time: a chain-of-thought reasoning directive and a fairness rule set.

For the reasoning directive, the VLM is asked to produce its reasoning in a \texttt{<thinking>} block before outputting its final label in a \texttt{<classification>} block (``Depressed'' or ``Not Depressed''). This encourages the model to ground its prediction in the clinical biomarkers in $\mathcal{B}_i$, and also gives us a reasoning trace that we can later score with the $\Delta REF$ audit.
For the fairness rule set, we append a short set of instructions that tell the model to base its assessment only on clinical indicators, ignore demographic factors such as gender and race, apply consistent standards across groups, and self-correct if it notices itself making demographic assumptions.
The full E-DAIC prompt templates, both baseline and intervention, are given in Appendix~\ref{app:prompts}.

\subsubsection{Counterfactual Fairness Algorithm (CFA)}
For the embedding-classification pipeline (Section \ref{sec:embedding_pipeline}), we made a training-time intervention to identify and penalise features that contribute to biased predictions, inspired by the framework used by Zhao \textit{et al.}~\cite{zhao2023fairness}. We achieve this by computing gradient-based feature attributions using integrated gradients~\cite{sundararajan2017axiomatic} for each feature during training.
If a feature has high importance for the prediction but is also strongly correlated with the sensitive attribute (e.g. gender), it is penalised. We modify the loss function to include this fairness penalty:
\begin{equation}
    \mathcal{L}_{\text{total}} = \mathcal{L}_{\text{CE}} + \lambda \mathcal{L}_{\text{CFA}}
\end{equation}
where $\lambda$ controls the trade-off between accuracy and fairness. We compared our baseline model against models trained with fairness weights $\lambda \in \{0.1, 0.5\}$ to observe the impact on fairness vs.\ accuracy.

\section{Experiments}
\subsection{Datasets}
We use two datasets; both datasets are in English and cover similar wellbeing contexts and are described below.

\vspace{1mm}
\noindent
\textbf{AFAR-BSFT Dataset}
consists of 44 weekly 20-minute sessions from 11 participants~\cite{cheong2023s}, where data was recorded in a dyadic setting with a well-being coach.
Extracted features include 709 facial features (Action Units, gaze, pose), 172 audio features, and two verbal biomarkers (label and probability).
AFAR-BSFT labels derive from participants' self-reported PANAS affect, in contrast to E-DAIC's PHQ-8. Both are psychometrically validated instruments, and PANAS Positive Affect is a well-established negative correlate of depression~\cite{watson1988positive}, which supports comparing patterns across the two datasets even though their labels are never pooled.

\vspace{1mm}
\noindent
\textbf{E-DAIC Dataset}
contains 275 semi-clinical interviews 
~\cite{ringeval2019avec}.
It included 73 hours of interviews, with the ground truth label being PHQ-8 scores.
Features comprise 17 facial Action Units via OpenFace, acoustic descriptors (pitch, loudness, voice jitter) and automatic text transcriptions.

\vspace{1mm}
\noindent
The two datasets required different preprocessing because they are distributed differently.
For \textbf{AFAR-BSFT}, the audio, visual and verbal biomarkers were supplied as separate per-session tables.
These were merged into a single per-session multimodal feature vector by joining on the session filename column, which uniquely identifies a participant–session pair.
In addition, sessions with missing values were removed.
For \textbf{E-DAIC}, each participant directory contained modality-specific files
keyed by participant ID rather than a shared filename. Frame-level audio and facial streams were collapsed to a single per-session vector by concatenating five summary statistics (mean, standard deviation, minimum, maximum, median) over each channel.
Missing or invalid entries within a frame were imputed to zero rather than removed, and a participant was excluded only when an entire modality file was missing.

\subsection{Model Selection}
We evaluate two state-of-the-art, small-scale VLMs that are suitable for resource-constrained environments (Google Colab T4 GPU).
The first is \textit{Qwen2-VL-2B-Instruct}, a 2 billion parameter model that introduces Naive Dynamic Resolution and Multimodal Rotary Position Embedding to enhance image and visual understanding that aligns closely with human perceptual processes~\cite{wang2024qwen2}.
The second is Microsoft's \textit{Phi-3.5-Vision-Instruct}, a 4.2 billion parameter model that is lightweight and built upon datasets that include high-quality synthetic data and filtered publicly available websites, ensuring reasoning-dense data across text and vision~\cite{abdin2024phi}.
%

\subsection{Evaluation}
Let $y$ be the ground truth, $\hat{y}$ be the model prediction, $TP$ represent true positives, $TN$ the true negatives, $FP$ the false positives and $FN$ the false negatives.
We report accuracy, F1-Score and Cohen's Kappa ($\kappa$) to account for potential class imbalances in the clinical labels, as seen in~\cite{carvalho2019machine}:

$$
\text{accuracy} = \frac{TP+TN}{TP+TN+FP+FN}
$$
$$
\text{F1-Score} = \frac{2 \times Precision \times Recall}{Precision + Recall}
$$
$$
\text{Precision} = \frac{TP}{TP+FP}, \ \text{Recall} = \frac{TP}{TP+FN}
$$

\subsubsection{Fairness Measures}
\label{sec:fairness_measures}
Let $A \in \{0,1\}$ be a sensitive attribute, where $A=1$ represents the majority group and $A=0$ represents the minority or protected group.
For example, in the AFAR-BSFT dataset, $A_{race}=1$ if the participant is white and $A_{race}=0$ if they are not white.
We evaluate the models using the following metrics, similar to existing works~\cite{yan2020mitigating}:

\begin{itemize}
    \item \textbf{Equal Accuracy} (EA), a group-based metric, can be calculated to compare group fairness between models. In other words, it is the absolute difference in classification accuracy between the majority and minority groups, where a high gap can indicate that the model is better calibrated for the majority group~\cite{yan2020mitigating}.
    $$
    EA = |P(\hat{y}=y | A=1) - P(\hat{y}=y | A=0)|
    $$
    
    \item \textbf{Equal opportunity} (EO), measures the true positive rate across all groups. Low true positive rates for an under-represented group could have serious consequences for diagnosis and treatment options~\cite{hardt2016equality}.
    $$
    EO = |P(\hat{y}=1 | y=1, A=1) - P(\hat{y}=1 | y=1, A=0)|
    $$
    
    \item \textbf{Disparate impact} (DI), measures whether positive outcomes occur at similar rates across groups, flagging where systems may not recognise positive outcomes for certain minority groups~\cite{feldman2015certifying}:
    $$
    DI = \frac{P(\hat{y}=1 | A=0)}{P(\hat{y}=1 | A=1)}
    $$    
    
\end{itemize}

\begin{table*}[t!]
\centering
\caption{\small Model performance and fairness measures for Qwen2-VL and Phi-3.5-Vision across AFAR-BSFT and E-DAIC Datasets. Best results denoted in \textbf{bold}.}
\vspace{-1em}
\label{tab:cross_dataset_results}
\setlength{\tabcolsep}{4pt}
\begin{tabular}{@{}lcccccccc@{}}
\toprule
 & \multicolumn{4}{c}{\textbf{AFAR-BSFT (Laboratory Dataset)}} & \multicolumn{4}{c}{\textbf{E-DAIC (In-the-Wild Dataset)}} \\
 \cmidrule(lr){2-5} \cmidrule(l{1em}){6-9}
 & \multicolumn{2}{c}{\textbf{Qwen2-VL}} & \multicolumn{2}{c}{\textbf{Phi-3.5-Vision}} & \multicolumn{2}{c}{\textbf{Qwen2-VL}} & \multicolumn{2}{c}{\textbf{Phi-3.5-Vision}} \\
 \cmidrule(lr){2-3} \cmidrule(lr){4-5} \cmidrule(lr){6-7} \cmidrule(l{1em}){8-9}
\textbf{Metric} & \textbf{Baseline} & \textbf{Intervention} & \textbf{Baseline} & \textbf{Intervention} & \textbf{Baseline} & \textbf{Intervention} & \textbf{Baseline} & \textbf{Intervention} \\
\midrule

\textit{Classification Performance} & & & & & & & & \\
Accuracy & \textbf{0.585} & 0.529 & 0.561 & 0.512 & 0.339 & 0.321 & \textbf{0.804} & 0.679 \\
Balanced Acc. & 0.509 & \textbf{0.589} & 0.522 & 0.480 & 0.526 & 0.513 & \textbf{0.743} & 0.720 \\
F1-Score & \textbf{0.730} & 0.385 & 0.667 & 0.615 & 0.479 & 0.472 & \textbf{0.645} & 0.609 \\
Miss Rate (FNR) & \textbf{0.042} & 0.750 & 0.250 & 0.333 & \textbf{0.000} & \textbf{0.000} & 0.412 & 0.177 \\
False Discovery Rate (FDR) & 0.400 & \textbf{0.167} & 0.294 & 0.429 & 0.685 & 0.691 & \textbf{0.286} & 0.517 \\
Cohen's Kappa ($\kappa$) & 0.020 & \textbf{0.155} & 0.047 & -0.041 & 0.032 & 0.016 & \textbf{0.511} & 0.366 \\
TN  & 1 & \textbf{13} & 5 & 5 & 2 & 1 & \textbf{35} & 24 \\
FP  & 16 & \textbf{1} & 12 & 12 & 37 & 38 & \textbf{4} & 15 \\
FN  & \textbf{1} & 15 & 6 & 8 & \textbf{0} & \textbf{0} & 7 & 3 \\
TP & \textbf{23} & 5 & 18 & 16 & \textbf{17} & \textbf{17} & 10 & 14 \\
\midrule

\textit{Fairness Audit: Gender} & & & & & & & & \\
$\Delta$ Accuracy & 0.196 & \textbf{0.083} & 0.093 & 0.198 & 0.360 & 0.383 & 0.145 & \textbf{0.118} \\
$\Delta$ Eq. Opportunity & \textbf{0.071} & 0.417 & 0.086 & 0.229 & \textbf{0.000} & \textbf{0.000} & 0.069 & 0.097 \\
Disp. Impact Ratio & \textbf{0.926} & 0.109 & 0.896 & 0.801 & 0.954 & \textbf{0.977} & 0.403 & 0.672 \\
$\Delta REF$ & 0.997 & 0.985 & 0.964 & \textbf{0.519} & — & — & — & — \\
\midrule
\textit{Fairness Audit: Race} & & & & & & & & \\
$\Delta$ Accuracy & 0.121 & 0.338 & 0.149 & \textbf{0.017} & — & — & — & — \\
$\Delta$ Eq. Opportunity & 0.056 & 0.467 & 0.333 & \textbf{0.000} & — & — & — & — \\
Disp. Impact Ratio & 1.074 & 2.778 & 1.208 & \textbf{0.967} & — & — & — & — \\
$\Delta REF$ & 0.960 & 0.926 & 0.940 & \textbf{0.414} & — & — & — & — \\
\bottomrule
\multicolumn{9}{l}{\footnotesize $^\dagger$Race metrics unavailable for E-DAIC (dataset lacks race annotations).}\\
\multicolumn{9}{l}{\footnotesize $^\ddagger$E-DAIC $\Delta$REF unavailable: outputs lack structured rationales required for explanation quality scoring.}
\end{tabular}%
\vspace{-1em}
\end{table*}

\subsubsection{Explainability Measures}
REF is a result-oriented metric and serves as a diagnostic tool to evaluate the consistency of a model's underlying logic ~\cite{zhao2023fairness}. It measures the disparity in Explanation Quality (EQ) between a minority group ($s=0$) and a majority group ($s=1$):
$$
\Delta REF = |P(\hat{q} = 1| s = 0) - P(\hat{q} = 1|s = 1)|
$$
where $\hat{q}$ is the quality label for the explanation.
An explanation is labelled as high-quality ($\hat{q}=1$) if the VLM successfully grounds its prediction in specific, valid clinical biomarkers provided in the prompt.
Otherwise, an explanation that is labelled low-quality ($\hat{q}=0$), exhibits ``hallucinated" rationales, relying on demographic proxies, or produces inconsistent reasoning that contradicts provided feature data.
The metric $\Delta REF$ measures the unfairness of high-quality proportion for subgroups, where a smaller value relates to a fairer model~\cite{zhao2023fairness}.

\subsection{Settings}
\noindent
\textbf{RQ1 Experiments: Zero-shot performance}
We deploy the generative classification pipeline, as explained in Section~\ref{sec:generative_pipeline}, across both the AFAR-BSFT and E-DAIC datasets.
We calculate and report standard classification metrics, including accuracy, F1-Score, precision and recall for both VLM models and datasets.

\vspace{1mm}
\noindent
\textbf{RQ2 Experiments: Performance disparities.}
We report fairness audit metrics, including Equal Accuracy (EA), Equal Opportunity (EO) and Disparate Impact (DI).
Additionally, we deploy the embedding-based classification pipeline, as explained in Section \ref{sec:embedding_pipeline}, on the E-DAIC dataset to perform a modality ablation suite, due to the raw modality data available.
We evaluate seven different configurations by extracting embeddings from the frozen VLM, ranging from unimodal (text, audio or facial only) to full-fusion, so we can observe how different modality signals may impact bias and performance.
%

\vspace{1mm}
\noindent
\textbf{RQ3 Experiments: XAI-guided interventions.}
We deploy our instructive fairness prompting strategy to the generative pipeline on both datasets, explicitly directing the VLMs to mitigate demographic bias, such as race and gender.
On the E-DAIC dataset, we implement the Counterfactual Fairness Algorithm (CFA) within the embedding pipeline, using integrated gradients for feature attribution during training.
Post-intervention, we recalculate all performance and fairness metrics and report Ratio-based Explanation Fairness ($\Delta REF$) across both datasets and VLM architectures to determine if these intervention strategies improve procedural consistency and reasoning quality across demographic subgroups.

\section{Results}

\label{results}
%
%

\noindent
\textbf{RQ1 Results (Zero-shot Evaluations):} across laboratory (AFAR-BSFT) and naturalistic (E-DAIC) domains revealed significant architectural discrepancies (Table~\ref{tab:cross_dataset_results}, ``Classification Performance'' rows).
On AFAR-BSFT, Qwen2-VL achieved a raw baseline accuracy of \textbf{0.585} and an F1-score of 0.730, while Phi-3.5-Vision yielded an accuracy of \textbf{0.561} and an F1-score of 0.667.
Additionally, Qwen2-VL displayed a Cohen's Kappa score of \textbf{$\kappa=0.020$}, while Phi-3.5-Vision was \textbf{$\kappa=0.047$}.
Of these, Qwen2-VL produced 16 false positives (1 false negative) and Phi-3.5-Vision produced 12 false positives (6 false negatives), with both models exhibiting a sensitivity-over-specificity profile (Table~\ref{tab:cross_dataset_results}, TN/FP/FN/TP rows).
For E-DAIC, under the zero-shot generative pipeline, Phi-3.5-Vision achieved \textbf{0.786} accuracy on text-only and improved to \textbf{0.804} in the full-fusion configuration.
In contrast, Qwen2-VL achieved only 0.482 accuracy on text-only, with full-fusion degrading performance to 0.339.

\vspace{2mm}
\noindent
\textbf{RQ2 Results (Performance disparities):}
Pre-intervention audits found that bias was present along both demographic axes for both models on AFAR-BSFT (Table~\ref{tab:cross_dataset_results}, ``Fairness Audit'' rows).
Comparing male and non-male groups, Qwen2-VL displayed an accuracy gap ($\Delta Acc$) of 0.196, a difference in opportunity ($\Delta EO$) of 0.071 and a disparate impact ratio ($DI$) of 0.926.
For Phi-3.5-Vision, we report a $\Delta Acc$ of 0.093, a $\Delta EO$ of 0.086 and $DI$ of 0.896.
Disparities more pronounced in race for Phi-3.5-Vision, exhibiting a baseline $\Delta Acc$ between white and non-white minority groups of 0.149, a $\Delta EO$ of 0.333 and $DI$ of 1.208.
Qwen2-VL showed lower initial racial disparities, with a $\Delta Acc$ of 0.121 and $\Delta EO$ of 0.056.
On E-DAIC, Phi-3.5-Vision under the generative pipeline exhibited relatively \textbf{low gender bias} ($\Delta EO=0.069$) across both text-only and full-fusion configurations (Table~\ref{tab:cross_dataset_results}).

\begin{figure}[t!]
    \centering
    \includegraphics[width=0.48\textwidth]{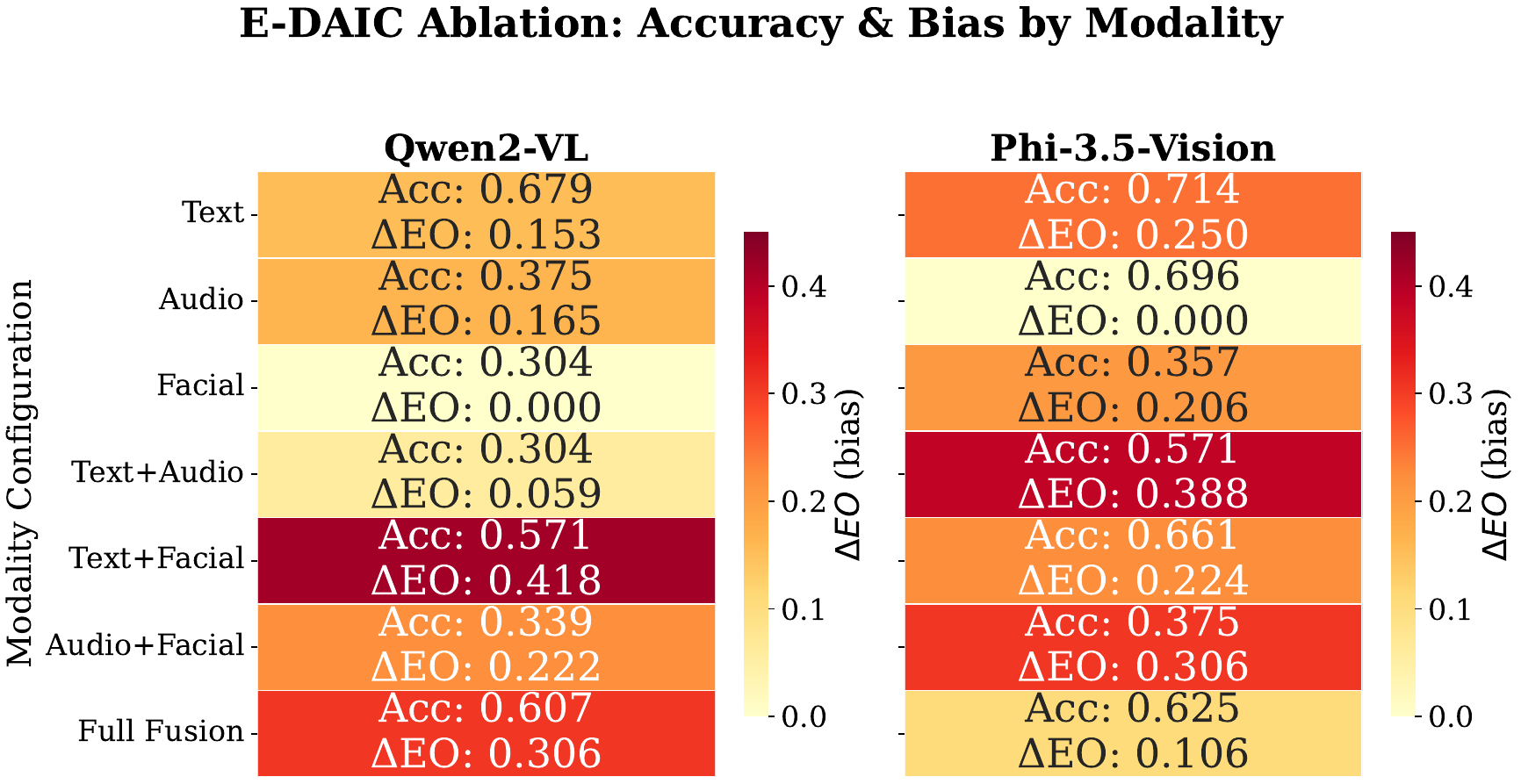}
    \caption{\small Heatmap of accuracy and $\Delta EO$ across all seven modality ablation configurations for Qwen2-VL and Phi-3.5-Vision on E-DAIC. Colour encodes $\Delta EO$ (darker = higher bias); cell text shows accuracy.}
    \label{fig:edaic_ablation}
    \Description{Heatmap depicting accuracy and bias results for each modality.}
        \vspace{-4mm}
\end{figure}

For E-DAIC, we also trained fusion classifiers across seven modality configurations, including \textit{text}, \textit{audio}, \textit{facial}, \textit{text and audio}, \textit{text and facial}, \textit{audio and facial}, in addition to \textit{full-fusion} , summarised in Table~\ref{tab:edaic_ablation} and visualised in Figure~\ref{fig:edaic_ablation}.
For Qwen2-VL, the \textit{text only} configuration achieved the \textbf{highest accuracy (0.679)} with moderate fairness metrics ($\Delta EO=0.153$). 
Incorporating additional modalities in the full fusion configuration worsened accuracy to 0.607 while increasing bias (\textbf{$\Delta EO=0.306$}).
Phi-3.5-Vision was similar, where \textit{text only} achieved \textbf{0.714} accuracy, though with higher bias ($\Delta EO=0.250$).
Full fusion achieved 0.625 accuracy with the \textbf{lowest bias observed} ($\Delta EO=0.106$).
The text and facial configuration showed the \textbf{highest bias amplification} for Qwen2-VL (\textbf{$\Delta EO=0.418$}).
The generative pipeline achieved $\Delta EO=0.069$ for Phi-3.5-Vision text-only on E-DAIC, while the embedding pipeline's text-only configuration recorded $\Delta EO=0.250$ for the same model, a notable disparity in equal opportunity between the two architectures (Table~\ref{tab:cross_dataset_results} and Table~\ref{tab:edaic_ablation}).

\begin{table}[t!]
\centering
\caption{\small E-DAIC Embedding-Based Results (Ablation Suite). Best results denoted in bold.}
\vspace{-1em}
\label{tab:edaic_ablation}
\begin{tabular}{lcccc}
\toprule
\textbf{Config} & \textbf{Qwen Acc} & \textbf{Qwen $\Delta$EO} & \textbf{Phi Acc} & \textbf{Phi $\Delta$EO} \\
\midrule
text\_only & \textbf{0.679} & 0.153 & \textbf{0.714} & 0.250 \\
audio\_only & 0.375 & 0.165 & 0.696 & 0.000* \\
facial\_only & 0.304 & 0.000* & 0.357 & 0.206 \\
text\_audio & 0.304 & \textbf{0.059} & 0.571 & 0.388 \\
text\_facial & 0.571 & 0.418 & 0.661 & 0.224 \\
audio\_facial & 0.339 & 0.222 & 0.375 & 0.306 \\
full\_fusion & 0.607 & 0.306 & 0.625 & \textbf{0.106} \\
\bottomrule
\multicolumn{5}{l}{\footnotesize *Zero bias due to model collapse (predicting single class)}
\end{tabular}
\vspace{-2em}
\end{table}

\vspace{2mm}
\noindent
\textbf{RQ3 Results (XAI-guided interventions):}
resulted in a decrease in accuracy for both models on AFAR-BSFT (Table~\ref{tab:cross_dataset_results}, ``Classification Performance'' rows under the AFAR-BSFT ``Intervention'' columns; corresponding confusion matrices in Figure~\ref{fig:afar_confusion}).
For Qwen2-VL, accuracy decreased to 0.529, and for Phi-3.5-Vision, it decreased to 0.512.
F1 decreased for both models to 0.385 and 0.615, while Cohen's Kappa increased to \textbf{$\kappa=0.155$} and decreased to \textbf{$\kappa=-0.041$} respectively.
Turning to fairness (Table~\ref{tab:cross_dataset_results}, ``Fairness Audit: Gender'' rows, Qwen2-VL ``Intervention'' column; visualised in Figure~\ref{fig:afar_radar}), balanced accuracy for Qwen2-VL increased from 0.509 to \textbf{0.589}, and gender disparities improved on $\Delta Acc$ ($0.196 \to 0.083$), though $DI$ moved further from parity ($0.926 \to 0.109$) and $\Delta EO$ increased to 0.417.
%
Across race (Table~\ref{tab:cross_dataset_results}, ``Fairness Audit: Race'' rows, Qwen2-VL ``Intervention'' column),
$\Delta Acc$ increased to 0.338, $\Delta EO$ rose to 0.467 and $DI$ increased to 2.778.
For Phi-3.5-Vision (Table~\ref{tab:cross_dataset_results}, ``Fairness Audit: Gender'' rows, Phi-3.5-Vision ``Intervention'' column; Figure~\ref{fig:afar_radar}), $\Delta Acc$ and $\Delta EO$ increased to 0.198 and 0.229, respectively, while $DI$ decreased to 0.801 for gender disparities (same gender rows, Phi-3.5-Vision columns).
In contrast, racial disparities saw an improvement (Table~\ref{tab:cross_dataset_results}, ``Fairness Audit: Race'' rows, Phi-3.5-Vision ``Intervention'' column), with $\Delta Acc$ decreasing to 0.017, $\Delta EO$ dropping to 0.000, and $DI$ moving closer to parity ($1.208 \to 0.967$).

For E-DAIC, the intervention on Qwen2-VL had \textbf{perfect equal opportunity} (\textbf{$\Delta EO=0.000$}) but with massively reduced accuracy (0.321) (Table~\ref{tab:cross_dataset_results}, E-DAIC Qwen2-VL ``Intervention'' column).
For Phi-3.5-Vision, post-intervention achieved better performance-fairness balance, maintaining \textbf{67.9\%} accuracy with $\Delta EO$ at 0.097 (Table~\ref{tab:cross_dataset_results}, E-DAIC Phi-3.5-Vision ``Intervention'' column).
We also performed a limited Pareto sweep for the Counterfactual Fairness Algorithm (CFA) with $\lambda \in \{0.1, 0.5\}$ to evaluate its impact on embedding-based fusion fairness (extending the baseline ablations of Table~\ref{tab:edaic_ablation}).
%
We had systematic model collapse across all tested configurations for both Qwen2-VL and Phi-3.5-Vision, where models converged to predicting the non-depressed class for all samples (Accuracy \textbf{0.696}, F1 0.000).
While this minimised fairness gaps ($\Delta EO = 0.000$), it made the models useless.

\section{Discussion}

\noindent \textbf{RQ1: How does zero-shot VLM depression classification performance vary between controlled laboratory datasets and naturalistic  clinical environments?}

Both VLM models demonstrate that in controlled laboratory environments (AFAR-BSFT), there is a tendency towards high sensitivity at the expense of specificity.
The high false-positive counts reported suggest that in the absence of fine-tuning, these baseline zero-shot models default to a conservative screening posture, prioritising the capture of depressive cues backed by high recall, even when cues may not indicate as such~\cite{guo2024bias,hu2023zero}.
%
This ensures vulnerable individuals are not overlooked~\cite{maxim2014screening}.
%
%
%
Furthermore, while the accuracy of 0.585 and 0.561 for Qwen2-VL and Phi-3.5-Vision represents our baseline capability of zero-shot reasoning, they remain notably inferior to state-of-the art multimodal models ~\cite{cameron2024multimodal}.
%

\begin{figure}[t!]
    \centering
    \includegraphics[width=0.48\textwidth]{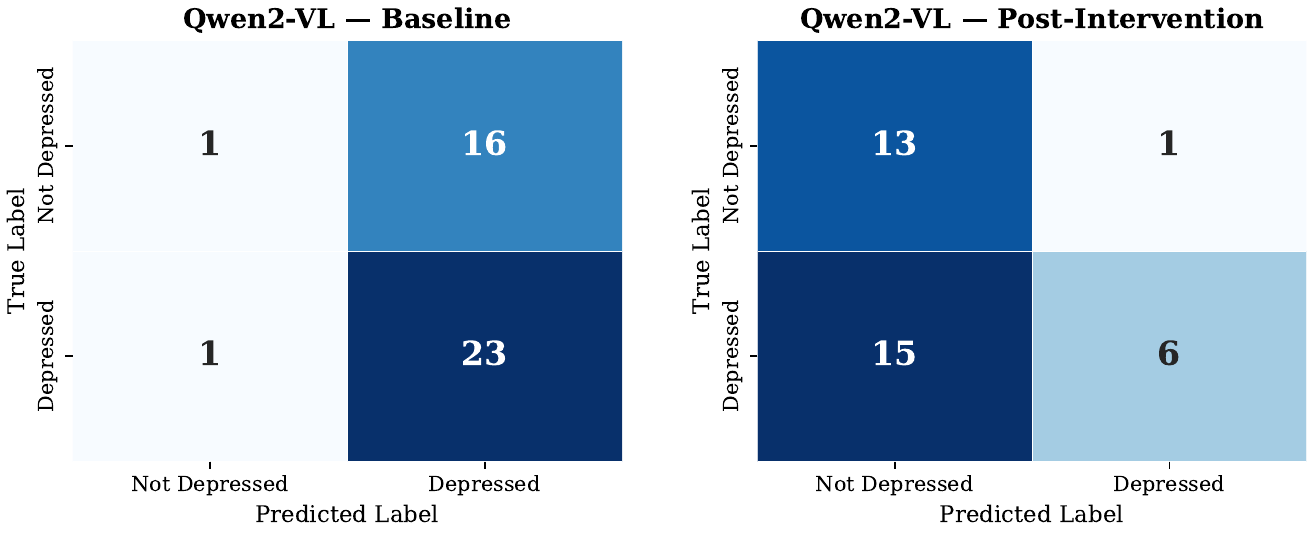}
    \hfill
    \includegraphics[width=0.48\textwidth]{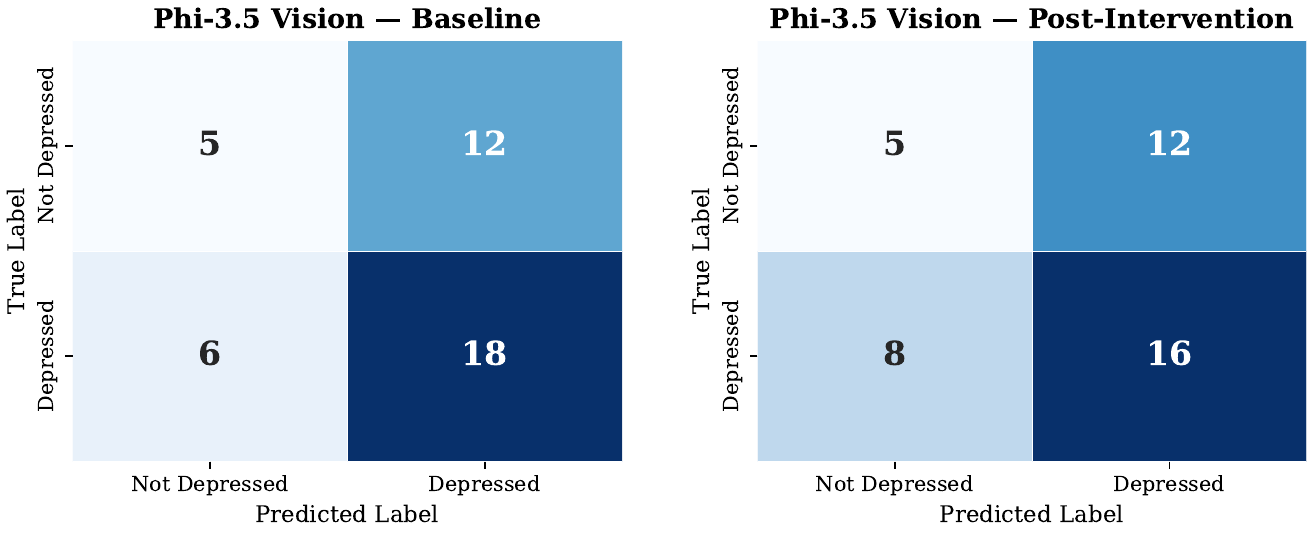}
    \caption{\small Confusion matrices for Qwen2-VL (left) and Phi-3.5-Vision (right) on the
    AFAR-BSFT dataset, before and after fairness intervention.}
    \label{fig:afar_confusion}
    \Description{Figure depicting confusion matrices for Qwen2-VL and Phi-3.5-Vision on the AFAR-BSFT dataset, before and after fairness intervention.}
        \vspace{-2mm}
\end{figure}

\vspace{2mm}
\noindent \textbf{(RQ2): To what extent do systematic performance disparities (e.g. gender-based accuracy) persist across different VLM architectures, and are these biases consistent across both controlled and naturalistic data domains?}

%
For the controlled laboratory AFAR-BSFT dataset, 
baseline disparities indicate that bias is present across both tested architectures. 
The AFAR-BSFT fairness audit (Section~\ref{results}, Table~\ref{tab:cross_dataset_results}; 
Figure~\ref{fig:afar_radar}) shows Qwen2-VL's bias 
is gender-dominated while Phi-3.5-Vision's is race-dominated. 
We interpret this divergence as architecture-specific: in the smaller Qwen2-VL, gendered facial cues appear to act as a stronger proxy for diagnostic decisions, whereas Phi-3.5-Vision is more prone to representation bias along racial lines, perhaps inherited from its training distribution. These AFAR-BSFT profiles rely on 44 sessions from 11 participants, and because repeated sessions from the same participant are not independent observations, we read them as indicative rather than precise.

On the E-DAIC dataset, bias values were heavily shaped by the chosen fusion strategy and classification paradigm.
%
%
The cross-paradigm comparison reported in Section~\ref{results} suggests that the generative pipeline is markedly fairer than the embedding pipeline at the same modality slice. 
We further note that adding multimodal signals dilutes text-specific bias under both paradigms, although the embedding approach has higher overall bias.
A possible explanation for the dilution effect 
is that text encoders in VLMs encode demographic proxies, like gendered language or dialect features, that can cause bias in the model and are spuriously correlated with the sensitive attribute~\cite{wolfe2022markedness, blodgett2016demographic}.
When audio and facial streams are incorporated, the fused representation is forced to integrate evidence from modalities that carry different, complementary diagnostic signals~\cite{poria2017review}. 
%
This effect was supported by audio-visual fusion systems exhibiting significantly lower demographic disparity than their unimodal counterparts~\cite{fenu2022demographic}.

\begin{figure}[t!]
    \centering
    \includegraphics[width=0.48\textwidth]{figures/afar_fairness_radar.pdf}
    \caption{\small Radar chart comparing fairness metrics ($\Delta$ Accuracy, $\Delta$
    Eq. Opportunity, Disparate Impact Ratio) across gender and race subgroups for both
    models on AFAR-BSFT, before and after intervention.}
    \label{fig:afar_radar}
    \Description{Figure depicts a radar chart comparing fairness metrics across gender and race subgroups for both models on AFAR-BSFT, before and after intervention.}
    \vspace{-4mm}
\end{figure}

\vspace{2mm}
\noindent \textbf{(RQ3): Can an XAI-guided intervention method be systematically deployed to improve VLM outcome fairness?}


In the controlled laboratory dataset (AFAR-BSFT), the intervention yielded a notable improvement in the diagnostic quality of Qwen2-VL: Cohen's $\kappa$ rose substantially and $\Delta REF$ decreased across all demographic subgroups (Figure~\ref{fig:afar_radar}), indicating more logically consistent and procedurally fair reasoning. 
The model's gender accuracy gap shrank but its opportunity difference and disparate impact ratio worsened, and racial bias was substantially amplified. We interpret this as evidence that forcing a model to ground its explanations may anchor it more firmly to biased internal representations of specific biomarkers, with XAI acting here as a diagnostic window into existing prejudices rather than as a corrective.

The impact on Phi-3.5-Vision presents a contrasting narrative: post-intervention, predictive validity collapses (Cohen's $\kappa$ becomes negative; see Figure~\ref{fig:afar_confusion}), and although racial fairness metrics improve dramatically, the simultaneous reduction in $\Delta REF$ must be interpreted with caution. When fairness gains coincide with sub-random performance, it is plausible that parity is achieved through random noise rather than equitable reasoning.
Such a result is functionally hollow and clinically unviable.
This finding underscores a broader ethical concern: explanations from VLMs derive their value from factual truthfulness, not surface plausibility. 

Table~\ref{tab:recommendations} consolidates the practical takeaways from this study, grouped by research question and pairing each finding with the recommendation it suggests.

\begin{table}[h]
\centering
\caption{\small Findings and recommendations from the FAIR\_XAI study, organised by research question.}
\vspace{-1em}
\label{tab:recommendations}
\small
\begin{tabular}{@{}p{0.05\columnwidth}p{0.38\columnwidth}p{0.45\columnwidth}@{}}
\toprule
\textbf{RQ} & \textbf{Findings} & \textbf{Recommendations} \\
\midrule
RQ1
  & Zero-shot VLM performance varies sharply across architectures and environments: Phi-3.5-Vision reached 80.4\% on E-DAIC while Qwen2-VL collapsed to 33.9\%. Both models over-predicted depression on AFAR-BSFT, with high false-positive rates and near-zero Cohen's $\kappa$.
  & Do not rely solely on controlled-lab benchmarks; require domain-invariant pre-training or fine-tuning before any deployment. Treat zero-shot VLM outputs as screening signals only and surface model uncertainty rather than confident labels. Clinicians must retain final diagnostic authority. \\
\addlinespace
RQ2
  & Bias is present in both architectures but is architecture-specific: Qwen2-VL is gender-dominated, Phi-3.5-Vision is race-dominated. Modality fusion can dilute or amplify bias depending on the combination -- \textit{text\,+\,facial} produced the worst observed disparity ($\Delta EO=0.418$).
  & Check fairness across multiple demographic axes simultaneously; single-axis checks will miss architecture-specific patterns. Report fairness for every modality combination, not only full-fusion, since intermediate fusions can amplify rather than average out bias. Restrict the underlying biometric features (facial units, voice, psychomotor signals) to research and clinical contexts. \\
\addlinespace
RQ3
  & XAI-guided interventions produced mixed outcomes. On AFAR-BSFT, Qwen2-VL gained procedural fairness ($\kappa$\,$\uparrow$, $\Delta REF$\,$\downarrow$) but its racial outcome bias was amplified. Phi-3.5-Vision's racial parity gains coincided with sub-random performance ($\kappa<0$), ``fairness through failure''. CFA caused systematic model collapse on E-DAIC.
  & Check explainability ($\Delta REF$) and outcome fairness ($\Delta EO$, $DI$) jointly, procedural transparency does not imply equitable outcomes. Reject any intervention whose fairness gain coincides with predictive collapse, and always report fairness alongside $\kappa$. Treat VLM rationales as material for independent verification against the prompt content, not as standalone justification, to guard against hallucinated explanations. \\
\bottomrule
\end{tabular}
\vspace{-1mm}
\end{table}

\section{Conclusion}

Our findings reveal key insights. First, zero-shot VLM performance varies substantially across environments and architectures. Phi-3.5-Vision had strong generative classification performance on the E-DAIC dataset (80.4\%), while Qwen2-VL struggled with the same task (33.9\%).
In addition, both models demonstrated a tendency to over-predict depression labels for the AFAR-BSFT dataset, showcasing fairly conservative screening assessments.
Secondly, demographic bias was present across both VLM architectures. Qwen2-VL showed higher gender-based disparities, whereas Phi-3.5-Vision had more racial bias. 
Thirdly, we audit two families of fairness intervention, both with mixed results. On E-DAIC, fairness prompting achieved perfect equal opportunity for Qwen2-VL at a severe accuracy cost. On AFAR-BSFT interventions improved procedural consistency but did not guarantee fairer outcomes, in some cases amplifying racial bias.
Our two datasets and two VLM architectures reflect the compute and data constraints of this study rather than a claim of broad generalisability across models or clinical domains. Further work could explore additional datasets and models for more robust benchmarking. In addition, training or fine-tuning a VLM that is domain invariant and able to recognise biomarkers despite the environmental setting might improve generalisability.
To further improve prediction bias, future work could explore how VLM adversarial training may affect model behaviour and whether a secondary independent model that judges whether rationales are demographically biased improves overall performance.
Drawing these threads together, Table~\ref{tab:recommendations} consolidates our findings into actionable recommendations for the responsible deployment of VLMs in clinical wellbeing assessment, organised by research question.

\section*{Safe and Responsible Innovation Statement}
This study acknowledges that automated systems in mental healthcare should augment, not replace, clinical judgement.
Risks include automation bias, where high-quality explanations or hallucinations might lead clinicians to overlook subtle diagnostic nuances or even disregard their own professional intuition, which could lead to misdiagnosis.
Therefore, transparency should serve as a mechanism for independent verification rather than just mere persuasion.
All participants in both datasets were anonymised, but the extraction of facial units, verbal features, and physical dynamics involves sensitive biometric data.
Such information should only be collected, processed, and stored within appropriate clinical or ethically approved research settings, with robust safeguards for privacy, security, and informed consent.
These models should not be repurposed 
for contexts that could result in misuse. 
Future deployment should also include human oversight, ongoing monitoring, regular auditing for fairness and performance across diverse populations, and clear governance mechanisms to ensure accountability and maintain public trust.





\newpage
\bibliographystyle{ACM-Reference-Format}
\bibliography{references}

@article{aung2015automatic,
  title={The automatic detection of chronic pain-related expression: requirements, challenges and the multimodal EmoPain dataset},
  author={Aung, Min SH and Kaltwang, Sebastian and Romera-Paredes, Bernardino and Martinez, Brais and Singh, Aneesha and Cella, Matteo and Valstar, Michel and Meng, Hongying and Kemp, Andrew and Shafizadeh, Moshen and others},
  journal={IEEE transactions on affective computing},
  volume={7},
  number={4},
  pages={435--451},
  year={2015},
  publisher={IEEE}
}

@inproceedings{cheong2026equitable,
  title={Equitable Robotics for Wellbeing (Eq-RW)},
  author={Cheong, Jiaee and Do{\u{g}}an, Fethiye Irmak and Markelius, Alva and Cross, Emily S and Eyssel, Friederike and Castellano, Ginevra and Gunes, Hatice},
  booktitle={Companion Proceedings of the 21st ACM/IEEE International Conference on Human-Robot Interaction},
  pages={1381--1383},
  year={2026}
}

@article{dwyer2025mindbench,
  title={Mindbench. ai: an actionable platform to evaluate the profile and performance of large language models in a mental healthcare context},
  author={Dwyer, Bridget and Flathers, Matthew and Sano, Akane and Dempsey, Allison and Cipriani, Andrea and Gazi, Asim H and Hill, Bryce and Gorban, Carla and Rodriguez, Carolyn I and Stromeyer IV, Charles and others},
  journal={NPP—Digital Psychiatry and Neuroscience},
  volume={3},
  number={1},
  pages={28},
  year={2025},
  publisher={Springer International Publishing Cham}
}

@article{cheong2026automated,
  title={Automated Detection Of Clinical High Risk Population Of Schizophrenia: Assessing The Generalizability Of NLP And LLM-Based Methods},
  author={Cheong, Jiaee and Corcoran, Cheryl M and Lewandowski, Kathryn E and Pasternak, Ofer and Kelly, Sinead and Bouix, Sylvain and Reichenberg, Abraham and Bearden, Carrie E and Cecchi, Guillermo and Baker, Justin T and others},
  year={2026}
}

@article{flathers2025interpreting,
  title={Interpreting psychiatric digital phenotyping data with large language models: a preliminary analysis},
  author={Flathers, Matthew and Xia, Winna and Hau, Christine and Nelson, Benjamin W and Cheong, Jiaee and Burns, James and Torous, John},
  journal={BMJ Mental Health},
  volume={28},
  number={1},
  year={2025},
  publisher={BMJ Publishing Group Ltd, Royal College of Psychiatrists and British~…}
}

@article{abbasi2025robot,
  title={Robot-Led Vision Language Model Wellbeing Assessment of Children},
  author={Abbasi, Nida Itrat and Dogan, Fethiye Irmak and Laban, Guy and Anderson, Joanna and Ford, Tamsin and Jones, Peter B and Gunes, Hatice},
  journal={arXiv preprint arXiv:2504.02765},
  year={2025}
}

@inproceedings{cheong2026fairssl,
  title={FairSSL: Fair Multimodal Self-Supervised Learning},
  author={Cheong, Jiaee and Mogharabin, Abtin and Liang, Paul Pu and Gunes, Hatice and Kalkan, Sinan},
  year={2026},
  booktitle={Forty-third International Conference on Machine Learning}
}

@article{chen2026current,
  title={The Current State/Trends in Digital Phenotyping for Mental Health Research and Care},
  author={Chen, Kelly and Torous, John and Cheong, Jiaee},
  journal={Psychiatric Clinics},
  volume={49},
  number={2},
  pages={273--287},
  year={2026},
  publisher={Elsevier}
}

@article{alghowinem2016multimodal,
  title={Multimodal depression detection: fusion analysis of paralinguistic, head pose and eye gaze behaviors},
  author={Alghowinem, Sharifa and Goecke, Roland and Wagner, Michael and Epps, Julien and Hyett, Matthew and Parker, Gordon and Breakspear, Michael},
  journal={IEEE Transactions on Affective Computing},
  volume={9},
  number={4},
  pages={478--490},
  year={2016},
  publisher={IEEE}
}

@inproceedings{laban2024lexi,
  title={Lexi: Large language models experimentation interface},
  author={Laban, Guy and Laban, Tomer and Gunes, Hatice},
  booktitle={Proceedings of the 12th international conference on human-agent interaction},
  pages={250--259},
  year={2024}
}

@inproceedings{dong2023towards,
  title={Towards next-generation intelligent assistants leveraging llm techniques},
  author={Dong, Xin Luna and Moon, Seungwhan and Xu, Yifan Ethan and Malik, Kshitiz and Yu, Zhou},
  booktitle={Proceedings of the 29th ACM SIGKDD Conference on Knowledge Discovery and Data Mining},
  pages={5792--5793},
  year={2023}
}

@article{laban2025people,
  title={What People Share With a Robot When Feeling Lonely and Stressed and How It Helps Over Time},
  author={Laban, Guy and Chiang, Sophie and Gunes, Hatice},
  journal={arXiv preprint arXiv:2504.02991},
  year={2025}
}

@inproceedings{cameron2024multimodal,
  title={Multimodal gender fairness in depression prediction: Insights on data from the usa \& china},
  author={Cameron, Joseph and Cheong, Jiaee and Spitale, Micol and Gunes, Hatice},
  booktitle={2024 12th International Conference on Affective Computing and Intelligent Interaction Workshops and Demos (ACIIW)},
  pages={265--273},
  year={2024},
  organization={IEEE}
}

@article{green2025gender,
  title={Gender Fairness of Machine Learning Algorithms for Pain Detection},
  author={Green, Dylan and Shang, Yuting and Cheong, Jiaee and Liu, Yang and Gunes, Hatice},
  journal={arXiv preprint arXiv:2506.11132},
  year={2025}
}

@inproceedings{lee2021included,
  title={Who is included in human perceptions of AI?: Trust and perceived fairness around healthcare AI and cultural mistrust},
  author={Lee, Min Kyung and Rich, Katherine},
  booktitle={Proceedings of the 2021 CHI conference on human factors in computing systems},
  pages={1--14},
  year={2021}
}

@article{helbling2025conceptattention,
  title={Conceptattention: Diffusion transformers learn highly interpretable features},
  author={Helbling, Alec and Meral, Tuna Han Salih and Hoover, Ben and Yanardag, Pinar and Chau, Duen Horng},
  journal={arXiv preprint arXiv:2502.04320},
  year={2025}
}

@inproceedings{zhao2023fairness,
  title={Fairness and explainability: Bridging the gap towards fair model explanations},
  author={Zhao, Yuying and Wang, Yu and Derr, Tyler},
  booktitle={Proceedings of the AAAI conference on artificial intelligence},
  volume={37},
  number={9},
  pages={11363--11371},
  year={2023}
}

@article{poria2017review,
  title={A review of affective computing: From unimodal analysis to multimodal fusion},
  author={Poria, Soujanya and Cambria, Erik and Bajpai, Rajiv and Hussain, Amir},
  journal={Information fusion},
  volume={37},
  pages={98--125},
  year={2017},
  publisher={Elsevier}
}

@inproceedings{lian2025mer,
  title={Mer 2025: When affective computing meets large language models},
  author={Lian, Zheng and Liu, Rui and Xu, Kele and Liu, Bin and Liu, Xuefei and Zhang, Yazhou and Liu, Xin and Li, Yong and Cheng, Zebang and Zuo, Haolin and others},
  booktitle={Proceedings of the 33rd ACM International Conference on Multimedia},
  pages={13837--13842},
  year={2025}
}

@article{chuang2023debiasing,
  title={Debiasing vision-language models via biased prompts},
  author={Chuang, Ching-Yao and Jampani, Varun and Li, Yuanzhen and Torralba, Antonio and Jegelka, Stefanie},
  journal={arXiv preprint arXiv:2302.00070},
  year={2023}
}

@inproceedings{radford2021learning,
  title={Learning transferable visual models from natural language supervision},
  author={Radford, Alec and Kim, Jong Wook and Hallacy, Chris and Ramesh, Aditya and Goh, Gabriel and Agarwal, Sandhini and Sastry, Girish and Askell, Amanda and Mishkin, Pamela and Clark, Jack and others},
  booktitle={International conference on machine learning},
  pages={8748--8763},
  year={2021},
  organization={PmLR}
}

@article{mehrabi2021survey,
  title={A survey on bias and fairness in machine learning},
  author={Mehrabi, Ninareh and Morstatter, Fred and Saxena, Nripsuta and Lerman, Kristina and Galstyan, Aram},
  journal={ACM computing surveys (CSUR)},
  volume={54},
  number={6},
  pages={1--35},
  year={2021},
  publisher={ACM New York, NY, USA}
}

@article{arrieta2020explainable,
  title={Explainable Artificial Intelligence (XAI): Concepts, taxonomies, opportunities and challenges toward responsible AI},
  author={Arrieta, Alejandro Barredo and D{\'\i}az-Rodr{\'\i}guez, Natalia and Del Ser, Javier and Bennetot, Adrien and Tabik, Siham and Barbado, Alberto and Garc{\'\i}a, Salvador and Gil-L{\'o}pez, Sergio and Molina, Daniel and Benjamins, Richard and others},
  journal={Information fusion},
  volume={58},
  pages={82--115},
  year={2020},
  publisher={Elsevier}
}

@inproceedings{lundberg2020explaining,
  title={Explaining quantitative measures of fairness},
  author={Lundberg, Scott M},
  booktitle={Fair \& Responsible AI Workshop@ CHI2020},
  year={2020}
}

@inproceedings{cheong2023s,
  title={“It’s not Fair!”--Fairness for a Small Dataset of Multi-modal Dyadic Mental Well-being Coaching},
  author={Cheong, Jiaee and Spitale, Micol and Gunes, Hatice},
  booktitle={2023 11th International Conference on Affective Computing and Intelligent Interaction (ACII)},
  pages={1--8},
  year={2023},
  organization={IEEE}
}

@inproceedings{ringeval2019avec,
  title={AVEC 2019 workshop and challenge: state-of-mind, detecting depression with AI, and cross-cultural affect recognition},
  author={Ringeval, Fabien and Schuller, Bj{\"o}rn and Valstar, Michel and Cummins, Nicholas and Cowie, Roddy and Tavabi, Leili and Schmitt, Maximilian and Alisamir, Sina and Amiriparian, Shahin and Messner, Eva-Maria and others},
  booktitle={Proceedings of the 9th International on Audio/visual Emotion Challenge and Workshop},
  pages={3--12},
  year={2019}
}

@article{wang2024qwen2,
  title={Qwen2-vl: Enhancing vision-language model's perception of the world at any resolution},
  author={Wang, Peng and Bai, Shuai and Tan, Sinan and Wang, Shijie and Fan, Zhihao and Bai, Jinze and Chen, Keqin and Liu, Xuejing and Wang, Jialin and Ge, Wenbin and others},
  journal={arXiv preprint arXiv:2409.12191},
  year={2024}
}

@article{abdin2024phi,
  title={Phi-4 technical report},
  author={Abdin, Marah and Aneja, Jyoti and Behl, Harkirat and Bubeck, S{\'e}bastien and Eldan, Ronen and Gunasekar, Suriya and Harrison, Michael and Hewett, Russell J and Javaheripi, Mojan and Kauffmann, Piero and others},
  journal={arXiv preprint arXiv:2412.08905},
  year={2024}
}

@inproceedings{yan2020mitigating,
  title={Mitigating biases in multimodal personality assessment},
  author={Yan, Shen and Huang, Di and Soleymani, Mohammad},
  booktitle={Proceedings of the 2020 international conference on multimodal interaction},
  pages={361--369},
  year={2020}
}

@article{carvalho2019machine,
  title={Machine learning interpretability: A survey on methods and metrics},
  author={Carvalho, Diogo V and Pereira, Eduardo M and Cardoso, Jaime S},
  journal={Electronics},
  volume={8},
  number={8},
  pages={832},
  year={2019},
  publisher={Multidisciplinary Digital Publishing Institute}
}

@inproceedings{sundararajan2017axiomatic,
  title     = {Axiomatic Attribution for Deep Networks},
  author    = {Sundararajan, Mukund and Taly, Ankur and Yan, Qiqi},
  booktitle = {Proceedings of the 34th International Conference on Machine Learning},
  pages     = {3319--3328},
  year      = {2017},
  publisher = {PMLR},
  url       = {https://proceedings.mlr.press/v70/sundararajan17a/sundararajan17a.pdf}
}

@inproceedings{hardt2016equality,
  title     = {Equality of Opportunity in Supervised Learning},
  author    = {Hardt, Moritz and Price, Eric and Srebro, Nati},
  booktitle = {Advances in Neural Information Processing Systems},
  volume    = {29},
  year      = {2016},
  url       = {https://arxiv.org/pdf/1610.02413}
}

@inproceedings{feldman2015certifying,
  title     = {Certifying and Removing Disparate Impact},
  author    = {Feldman, Michael and Friedler, Sorelle A. and Moeller, John and
               Scheidegger, Carlos and Venkatasubramanian, Suresh},
  booktitle = {Proceedings of the 21st ACM SIGKDD International Conference on
               Knowledge Discovery and Data Mining},
  pages     = {259--268},
  year      = {2015},
  url       = {https://arxiv.org/pdf/1412.3756}
}

@inproceedings{wolfe2022markedness,
  title={Markedness in Visual Semantic {AI}},
  author={Wolfe, Robert and Caliskan, Aylin},
  booktitle={Proceedings of the 2022 ACM Conference on Fairness, Accountability, and Transparency},
  pages={1266--1279},
  year={2022},
  publisher={ACM}
}

@inproceedings{blodgett2016demographic,
  title={Demographic Dialectal Variation in Social Media: A Case Study of {African-American English}},
  author={Blodgett, Su Lin and Green, Lisa and O{'}Connor, Brendan},
  booktitle={Proceedings of the 2016 Conference on Empirical Methods in Natural Language Processing},
  pages={1119--1130},
  year={2016},
  publisher={ACL}
}

@article{fenu2022demographic,
  title={Demographic Fairness in Multimodal Biometrics: A Comparative Analysis on
         Audio-Visual Speaker Recognition Systems},
  author={Fenu, Gianni and Marras, Mirko},
  journal={Procedia Computer Science},
  volume={198},
  pages={249--254},
  year={2022},
  publisher={Elsevier}
}

@article{domnich2021responsible,
  title={Responsible AI: Gender bias assessment in emotion recognition},
  author={Domnich, Artem and Anbarjafari, Gholamreza},
  journal={arXiv preprint arXiv:2103.11436},
  year={2021}
}

@inproceedings{taubenfeld2024systematic,
  title={Systematic biases in LLM simulations of debates},
  author={Taubenfeld, Amir and Dover, Yaniv and Reichart, Roi and Goldstein, Ariel},
  booktitle={Proceedings of the 2024 conference on empirical methods in natural language processing},
  pages={251--267},
  year={2024}
}

@article{guo2024bias,
  title={Bias in large language models: Origin, evaluation, and mitigation},
  author={Guo, Yufei and Guo, Muzhe and Su, Juntao and Yang, Zhou and Zhu, Mengqiu and Li, Hongfei and Qiu, Mengyang and Liu, Shuo Shuo},
  journal={arXiv preprint arXiv:2411.10915},
  year={2024}
}

@inproceedings{gui2019cooperative,
  title={Cooperative multimodal approach to depression detection in twitter},
  author={Gui, Tao and Zhu, Liang and Zhang, Qi and Peng, Minlong and Zhou, Xu and Ding, Keyu and Chen, Zhigang},
  booktitle={Proceedings of the AAAI conference on artificial intelligence},
  volume={33},
  number={01},
  pages={110--117},
  year={2019}
}

@article{dwivedi2023explainable,
  title={Explainable AI (XAI): Core ideas, techniques, and solutions},
  author={Dwivedi, Rudresh and Dave, Devam and Naik, Het and Singhal, Smiti and Omer, Rana and Patel, Pankesh and Qian, Bin and Wen, Zhenyu and Shah, Tejal and Morgan, Graham and others},
  journal={ACM computing surveys},
  volume={55},
  number={9},
  pages={1--33},
  year={2023},
  publisher={ACM New York, NY}
}

@article{cortinas2023toward,
  title={Toward explainable affective computing: A review},
  author={Corti{\~n}as-Lorenzo, Karina and Lacey, Gerard},
  journal={IEEE Transactions on Neural Networks and Learning Systems},
  volume={35},
  number={10},
  pages={13101--13121},
  year={2023},
  publisher={IEEE}
}

@article{nori2019interpretml,
  title={Interpretml: A unified framework for machine learning interpretability},
  author={Nori, Harsha and Jenkins, Samuel and Koch, Paul and Caruana, Rich},
  journal={arXiv preprint arXiv:1909.09223},
  year={2019}
}

@article{thalpage2023unlocking,
  title={Unlocking the black box: Explainable artificial intelligence (XAI) for trust and transparency in ai systems},
  author={Thalpage, Nipuna},
  journal={J. Digit. Art Humanit},
  volume={4},
  number={1},
  pages={31--36},
  year={2023}
}

@inproceedings{arias2022focus,
  title={Focus! rating xai methods and finding biases},
  author={Arias-Duart, Anna and Par{\'e}s, Ferran and Garcia-Gasulla, Dario and Gim{\'e}nez-{\'A}balos, Victor},
  booktitle={2022 IEEE International Conference on Fuzzy Systems (FUZZ-IEEE)},
  pages={1--8},
  year={2022},
  organization={IEEE}
}

@article{shannon1948mathematical,
  title={A mathematical theory of communication},
  author={Shannon, Claude Elwood},
  journal={The Bell system technical journal},
  volume={27},
  number={3},
  pages={379--423},
  year={1948},
  publisher={Nokia Bell Labs}
}

@article{roubhi2025mutual,
  title={Mutual information-based feature selection strategy for speech emotion recognition using machine learning algorithms combined with the voting rules method},
  author={Roubhi, Hamza and Gharbi, Abdenour Hacine and Rouabah, Khaled and Ravier, Philippe},
  journal={Engineering, Technology \& Applied Science Research},
  volume={15},
  number={1},
  pages={19207--19213},
  year={2025}
}

@article{amiri2011mutual,
  title={Mutual information-based feature selection for intrusion detection systems},
  author={Amiri, Fatemeh and Yousefi, MohammadMahdi Rezaei and Lucas, Caro and Shakery, Azadeh and Yazdani, Nasser},
  journal={Journal of network and computer applications},
  volume={34},
  number={4},
  pages={1184--1199},
  year={2011},
  publisher={Elsevier}
}

@article{hu2023zero,
  title={Zero-shot clinical entity recognition using chatgpt},
  author={Hu, Yan and Ameer, Iqra and Zuo, Xu and Peng, Xueqing and Zhou, Yujia and Li, Zehan and Li, Yiming and Li, Jianfu and Jiang, Xiaoqian and Xu, Hua},
  journal={arXiv preprint arXiv:2303.16416},
  year={2023},
  publisher={May}
}

@article{maxim2014screening,
  title={Screening tests: a review with examples},
  author={Maxim, L Daniel and Niebo, Ron and Utell, Mark J},
  journal={Inhalation toxicology},
  volume={26},
  number={13},
  pages={811--828},
  year={2014},
  publisher={Taylor \& Francis}
}

@inproceedings{cheong2023causal,
  title={Causal Structure Learning of Bias for Fair Affect Recognition},
  author={Cheong, Jiaee and Kalkan, Sinan and Gunes, Hatice},
  booktitle={Proceedings of the IEEE/CVF Winter Conference on Applications of Computer Vision},
  year={2023}
}

@article{cheong2022counterfactual,
  title={Counterfactual Fairness for Facial Expression Recognition},
  author={Cheong, Jiaee and Kalkan, Sinan and Gunes, Hatice},
  journal={ECCV Workshop on Challenge on People Analysis (WCPA)},
  year={2022}
}

@article{cheong2021hitchhiker,
  title={The hitchhiker’s guide to bias and fairness in facial affective signal processing: Overview and techniques},
  author={Cheong, Jiaee and Kalkan, Sinan and Gunes, Hatice},
  journal={IEEE Signal Processing Magazine},
  volume={38},
  number={6},
  pages={39--49},
  year={2021},
  publisher={IEEE}
}

@inproceedings{cheong_gender_fairness,
  title={Towards gender fairness for mental health prediction},
  author={Cheong, Jiaee and Kuzucu, Selim and Kalkan, Sinan and Gunes, Hatice},
  booktitle={Proceedings of the Thirty-Second International Joint Conference on Artificial Intelligence},
  pages={5932--5940},
  year={2023}
}

@article{spitale2024underneath,
  title={Underneath the Numbers: Quantitative and Qualitative Gender Fairness in LLMs for Depression Prediction},
  author={Spitale, Micol and Cheong, Jiaee and Gunes, Hatice},
  journal={arXiv preprint arXiv:2406.08183},
  year={2024}
}

@article{cheong2025small,
  title={Small but fair! fairness for multimodal human-human and robot-human mental wellbeing coaching},
  author={Cheong, Jiaee and Spitale, Micol and Gunes, Hatice},
  journal={IEEE Transactions on Affective Computing},
  year={2025},
  publisher={IEEE}
}

@inproceedings{cheong2025u,
  title={U-Fair: Uncertainty-based Multimodal Multitask Learning for Fairer Depression Detection},
  author={Cheong, Jiaee and Bangar, Aditya and Kalkan, Sinan and Gunes, Hatice},
  booktitle={Machine Learning for Health (ML4H)},
  pages={203--218},
  year={2025},
  organization={PMLR}
}

@inproceedings{kwok2025machine,
  title={Machine learning fairness for depression detection using eeg data},
  author={Kwok, Angus Man Ho and Cheong, Jiaee and Kalkan, Sinan and Gunes, Hatice},
  booktitle={2025 IEEE 22nd International Symposium on Biomedical Imaging (ISBI)},
  pages={1--5},
  year={2025},
  organization={IEEE}
}

@inproceedings{spitale2025exploring,
  title={Exploring Causality for HRI: A Case Study on Robotic Mental Well-being Coaching},
  author={Spitale, Micol and Babu, Srikar and Cakmak, Serhan and Cheong, Jiaee and Gunes, Hatice},
  booktitle={2025 34th IEEE International Conference on Robot and Human Interactive Communication (RO-MAN)},
  pages={708--713},
  year={2025},
  organization={IEEE}
}

@article{zhang2024vision,
  title={Vision-language models for vision tasks: A survey},
  author={Zhang, Jingyi and Huang, Jiaxing and Jin, Sheng and Lu, Shijian},
  journal={IEEE transactions on pattern analysis and machine intelligence},
  volume={46},
  number={8},
  pages={5625--5644},
  year={2024},
  publisher={IEEE}
}

@article{cheong2025fairwell,
  title={Fairwell: Fair multimodal self-supervised learning for wellbeing prediction},
  author={Cheong, Jiaee and Mogharabin, Abtin and Liang, Paul and Gunes, Hatice and Kalkan, Sinan},
  journal={arXiv preprint arXiv:2508.16748},
  year={2025}
}

@inproceedings{cheong2024fairrefuse,
  title={FairReFuse: referee-guided fusion for multimodal causal fairness in depression detection},
  author={Cheong, Jiaee and Kalkan, Sinan and Gunes, Hatice},
  booktitle={Proceedings of the Thirty-Third International Joint Conference on Artificial Intelligence},
  pages={7224--7232},
  year={2024}
}

@article{dogan2026investigating,
  title={Investigating Associational Biases in Inter-Model Communication of Large Generative Models},
  author={Dogan, Fethiye Irmak and Weiss, Yuval and Patel, Kajal and Cheong, Jiaee and Gunes, Hatice},
  journal={arXiv preprint arXiv:2601.22093},
  year={2026}
}

@article{ratti2022explainable,
  title={Explainable machine learning practices: opening another black box for reliable medical AI},
  author={Ratti, Emanuele and Graves, Mark},
  journal={AI and Ethics},
  volume={2},
  number={4},
  pages={801--814},
  year={2022},
  publisher={Springer}
}

@misc{who2025depression,
  title = {Depressive disorder (depression)},
  author={},
  year = {2025},
  howpublished = {\url{https://www.who.int/news-room/fact-sheets/detail/depression}},
  note = {Accessed: 2026-04-22}
}

@article{song2025automatic,
  title={Automatic Depression Assessment using Machine Learning: A Comprehensive Survey},
  author={Song, Siyang and Huo, Yupeng and Tang, Shiqing and Cheong, Jiaee and Gao, Rui and Valstar, Michel and Gunes, Hatice},
  journal={arXiv preprint arXiv:2506.18915},
  year={2025}
}

@article{li2025automated,
  title={Automated depression detection from text and audio: A systematic review},
  author={Li, Yuxin and Kumbale, Sinchana and Chen, Yanru and Surana, Tanmay and Chng, Eng Siong and Guan, Cuntai},
  journal={IEEE Journal of Biomedical and Health Informatics},
  year={2025},
  publisher={IEEE}
}

@article{watson1988positive,
  title={Positive and negative affectivity and their relation to anxiety and depressive disorders},
  author={Watson, David and Clark, Lee Anna and Carey, Gregory},
  journal={Journal of Abnormal Psychology},
  volume={97},
  number={3},
  pages={346--353},
  year={1988},
  publisher={American Psychological Association}
}

\clearpage
\appendix

\section{Appendix}
\label{app:prompts}
\subsection{E-DAIC Prompt Templates}

For completeness, we give the prompt templates used by the E-DAIC generative pipeline (Section~\ref{sec:generative_pipeline}). Following $\mathcal{P}_i = \mathcal{D}_i \oplus \mathcal{B}_i \oplus \mathcal{T}$, each participant prompt concatenates a demographic block $\mathcal{D}_i$, the top-$k$ biomarkers $\mathcal{B}_i$ selected by mutual information, and a fixed task instruction $\mathcal{T}$. Values in braces are substituted per participant. E-DAIC carries a gender annotation but no race label, so $\mathcal{D}_i$ holds gender alone; the biomarkers are the OpenFace Action Units and eGeMAPS acoustic descriptors summarised over the session.

\vspace{2mm}
\noindent
\textbf{Baseline prompt (RQ1--RQ2).} The zero-shot template, with no fairness intervention applied.

\begin{verbatim}
You are a clinical wellbeing assessment
assistant.

Participant demographics:
  Gender: {gender}

Clinical biomarkers (top-k by mutual
information):
  {biomarker_1}: {value_1}
  {biomarker_2}: {value_2}
  ...
  {biomarker_k}: {value_k}

Using only the biomarkers above, classify
this participant as "Depressed" or "Not
Depressed". Return the predicted label with
a confidence score and a brief justification.
\end{verbatim}
\normalsize

\vspace{2mm}
\noindent
\textbf{Fairness intervention (RQ3).} The demographic and biomarker blocks are unchanged. We append two components to the task instruction $\mathcal{T}$: a chain-of-thought directive and a fairness rule set. 

\begin{verbatim}
Work through your reasoning inside a
<thinking> block, grounded in the biomarkers
above. Then give your final label inside a
<classification> block as either "Depressed"
or "Not Depressed".

Fairness rules:
  1. Base your assessment only on the
     clinical biomarkers.
  2. Ignore demographic factors such as
     gender and race.
  3. Apply the same standard to every
     participant.
  4. If you notice a demographic assumption
     influencing you, correct it before
     deciding.
\end{verbatim}
\normalsize

\end{document}